\documentclass[conference]{IEEEtran}
\IEEEoverridecommandlockouts
\usepackage{bbold}
\usepackage{graphicx}
\usepackage{multirow}
\usepackage{subcaption}
\usepackage{enumitem}
\usepackage{soul}
\usepackage{color}

\usepackage{acronym}
\usepackage{xspace}
\usepackage[normalem]{ulem}

\usepackage{tabularx}
\usepackage{multirow}
\usepackage{booktabs}

\usepackage{comment}
\usepackage{amsmath,amssymb,amsfonts}
\usepackage{algorithmic}
\usepackage{graphicx}
\usepackage{textcomp}
\usepackage{xcolor}
\usepackage[table]{xcolor}
\usepackage{tikz}
\usepackage{wrapfig}
\usepackage[sort,space]{cite}
\usepackage[ruled,vlined]{algorithm2e}
\usepackage[usestackEOL]{stackengine}

\definecolor{ForestGreen}{RGB}{34,139,34}
\definecolor{NavyBlue}{RGB}{0,0,128}

\usepackage[font=bf]{caption}
\usepackage[T1]{fontenc}
\usepackage{fancyvrb}

\newcommand*\circled[1]{\tikz[baseline=(char.base)]{\node[shape=circle,draw,inner sep=1pt,font=\small] (char) {#1};}}

\SetKwInput{KwInput}{\textcolor{blue}{\textbf{Input}}}
\SetKwInput{KwOutput}{\textcolor{blue}{\textbf{Output}}}
\SetKwIF{If}{ElseIf}{Else}{\textcolor{blue}{\textbf{if}}}{\textcolor{blue}{\textbf{then}}}{\textcolor{blue}{\textbf{else if}}}{\textcolor{blue}{\textbf{else}}}{\textcolor{blue}{\textbf{endif}}}
\SetKwFor{For}{\textcolor{blue}{\textbf{for}}}{\textcolor{blue}{\textbf{do}}}{\textcolor{blue}
{\textbf{endfor}}}
\SetKwFunction{Length}{length}
\SetKwFunction{Max}{max}
\SetKwProg{Fn}{\textcolor{blue}{\textbf{function}}}{}{}
\SetKw{KwRet}{\textcolor{blue}{\textbf{return}}}

\newcommand{\sys}{\textit{pFedNavi}\xspace}

\def\BibTeX{{\rm B\kern-.05em{\sc i\kern-.025em b}\kern-.08em
    T\kern-.1667em\lower.7ex\hbox{E}\kern-.125emX}}

\begin{document}

\title{\sys: Structure-Aware Personalized Federated Vision-Language Navigation for Embodied AI
\thanks{This work was supported in part by the National Natural Science Foundation of China (NO. 62472284),  Openmind (Wuhu) Intelligent Robot Co., Ltd., and Shanghai Key Laboratory of Scalable Computing and Systems. M. Pan’s work was supported in part by the US National Science Foundation under grants CNS-2107057, CNS-2318664, CSR-2403249, and CNS-2431596.}
}

\author{
    \IEEEauthorblockN{\parbox[t]{0.3\textwidth}{\centering
        Qingqian~Yang, Hao~Wang}}
    \IEEEauthorblockA{\parbox[t]{0.3\textwidth}{\centering
        \textit{Stevens Institute of Technology}\\
        Hoboken, NJ, USA\\
        \texttt{\{qyang21,hwang9\}@stevens.edu}
    }}
    \and
    \IEEEauthorblockN{\parbox[t]{0.3\textwidth}{\centering
        Sai~Qian~Zhang}}
    \IEEEauthorblockA{\parbox[t]{0.3\textwidth}{\centering
        \textit{New York University}\\
        New York, NY, USA\\
        \texttt{sai.zhang@nyu.edu}
    }}
    \and
    \IEEEauthorblockN{\parbox[t]{0.3\textwidth}{\centering
        Jian~Li}}
    \IEEEauthorblockA{\parbox[t]{0.3\textwidth}{\centering
        \textit{Stony Brook University}\\
        Stony Brook, NY, USA\\
        \texttt{jian.li.3@stonybrook.edu}
    }}
    \and
    \IEEEauthorblockN{\parbox[t]{0.26\textwidth}{\centering
        Yang~Hua}}
    \IEEEauthorblockA{\parbox[t]{0.26\textwidth}{\centering
        \textit{Queen's University Belfast}\\
        Belfast, UK\\
        \texttt{Y.Hua@qub.ac.uk}
    }}
    \and
    \IEEEauthorblockN{\parbox[t]{0.3\textwidth}{\centering
        Miao~Pan}}
    \IEEEauthorblockA{\parbox[t]{0.3\textwidth}{\centering
        \textit{University of Houston}\\
        Houston, TX, USA\\
        \texttt{mpan2@central.uh.edu}
    }}
    \and
    \IEEEauthorblockN{\parbox[t]{0.4\textwidth}{\centering
        Tao~Song, Zhengwei~Qi, Haibing~Guan}}
    \IEEEauthorblockA{\parbox[t]{0.44\textwidth}{\centering
        \textit{Shanghai Jiao Tong University}\\
        Shanghai, China\\
        \texttt{\{songt333,qizhwei,hbguan\}@sjtu.edu.cn}
    }}
}

\maketitle

\begin{abstract}

Vision-Language Navigation (VLN) requires large-scale trajectory–instruction data from private indoor environments, raising significant privacy concerns. While Federated Learning (FL) mitigates this by keeping data on-device, \emph{vanilla} FL struggles under VLN’s extreme cross-client heterogeneity in environments and instruction styles, rendering a single global model suboptimal. 
This paper proposes \textbf{\sys}, a \emph{structure-aware} and \emph{dynamically adaptive} personalized federated learning framework tailored for VLN. 
Our key idea is to personalize \emph{where it matters}: \sys (i) adaptively identifies client-specific layers via layer-wise mixing coefficients, and (ii) performs fine-grained parameter fusion on the selected components (e.g., the encoder--decoder projection and environment-sensitive decoder layers) to balance global knowledge sharing with local specialization.
We evaluate \sys on two standard VLN benchmarks, R2R and RxR, using both ResNet and CLIP visual representations. Across all metrics, \sys consistently outperforms the FedAvg-based VLN baseline, achieving up to 7.5\% improvement in navigation success rate and up to 7.8\% gain in trajectory fidelity, while converging 1.38$\times$ faster under non-IID conditions.  

\end{abstract}


\acrodef{ML}[ML]{Machine Learning}
\acrodef{ML}[ML]{Machine Learning}
\newcommand{\ML}{\ac{ML}\xspace}

\acrodef{NSF}[NSF]{National Science Foundation}
\newcommand{\NSF}{\ac{NSF}\xspace}

\acrodef{AI}[AI]{Artificial Intelligence}
\newcommand{\AI}{\ac{AI}\xspace}

\acrodef{CIDA}[CIDA]{Continuously Indexed Domain Adaptation}
\newcommand{\CIDA}{\ac{CIDA}\xspace}

\acrodef{KS}[KS]{Kolmogorov-Smirnov}
\newcommand{\KS}{\ac{KS}\xspace}

\acrodef{PCA}[PCA]{Principal Component Analysis}
\newcommand{\PCA}{\ac{PCA}\xspace}

\acrodef{FL}[FL]{Federated Learning}
\newcommand{\FL}{\ac{FL}\xspace}

\acrodef{FDA}[FDA]{Federated Domain Adaptation}
\newcommand{\FDA}{\ac{FDA}\xspace}

\acrodef{CL}[CL]{Critical Learning}
\newcommand{\CL}{\ac{CL}\xspace}

\acrodef{AC}[AC]{Attacking-Critical}
\newcommand{\AC}{\ac{AC}\xspace}

\acrodef{CAGR}[CAGR]{compound annual growth rate}
\newcommand{\CAGR}{\ac{CAGR}\xspace}

\acrodef{CCT}[CCT]{Center for Computation and Technology}
\newcommand{\CCT}{\ac{CCT}\xspace}

\acrodef{SLO}[SLO]{service level objective}
\newcommand{\SLO}{\ac{SLO}\xspace}

\acrodefplural{SLOs}[SLO]{service level objectives}
\newcommand{\SLOs}{\acp{SLO}\xspace}

\acrodef{RL}[RL]{reinforcement learning}
\newcommand{\RL}{\ac{RL}\xspace}

\acrodef{DRL}[DRL]{deep reinforcement learning}
\newcommand{\DRL}{\ac{DRL}\xspace}

\acrodef{VM}[VM]{virtual machine}
\newcommand{\VM}{\ac{VM}\xspace}

\acrodefplural{VM}[VMs]{virtual machines}
\newcommand{\VMs}{\acp{VM}\xspace}

\acrodef{ITC}[ITC]{Innovation \& Technology Commercialization}
\newcommand{\ITC}{\ac{ITC}\xspace}

\acrodef{DAG}[DAG]{directed acyclic graph}
\newcommand{\DAG}{\ac{DAG}\xspace}

\acrodefplural{DAG}[DAGs]{directed acyclic graphs}
\newcommand{\DAGs}{\acp{DAG}\xspace}

\acrodef{SFA}[SFA]{single point authentication}
\newcommand{\SFA}{\ac{SFA}\xspace}

\acrodef{HPC}[HPC]{high-performance computing}
\newcommand{\HPC}{\ac{HPC}\xspace}

\acrodef{SBIR}[SBIR]{Small Business Innovation Research}
\newcommand{\SBIR}{\ac{SBIR}\xspace}

\acrodef{IoT}[IoT]{Internet of Things}
\newcommand{\IoT}{\ac{IoT}\xspace}

\acrodef{DML}[DML]{distributed machine learning}
\newcommand{\DML}{\ac{DML}\xspace}

\acrodef{GNN}[GNN]{graph neural network}
\newcommand{\GNN}{\ac{GNN}\xspace}

\acrodefplural{GNN}[GNNs]{graph neural networks}
\newcommand{\GNNs}{\acp{GNN}\xspace}

\acrodef{BSR}[BSR]{backdoor success rate}
\newcommand{\BSR}{\ac{BSR}\xspace}

\acrodef{BTA}[BTA]{backdoor task accuracy}
\newcommand{\BTA}{\ac{BTA}\xspace}

\acrodef{ATT}[ATT]{App Tracking Transparency}
\newcommand{\ATT}{\ac{ATT}\xspace}

\acrodef{DNN}[DNN]{deep neural network}
\newcommand{\DNN}{\ac{DNN}\xspace}

\acrodef{DP}[DP]{differential privacy}
\newcommand{\DP}{\ac{DP}\xspace}

\acrodef{CDA}[CDA]{centralized domain adaptation}
\newcommand{\CDA}{\ac{CDA}\xspace}

\acrodef{DIA}[DIA]{distribution inference attack}
\newcommand{\DIA}{\ac{DIA}\xspace}

\acrodef{MIA}[MIA]{membership inference attack}
\newcommand{\MIA}{\ac{MIA}\xspace}

\acrodef{MSE}[MSE]{mean square error}
\newcommand{\MSE}{\ac{MSE}\xspace}

\acrodef{IID}[IID]{independent and identically distributed}
\newcommand{\IID}{\ac{IID}\xspace}

\acrodef{SOTA}[SOTA]{state-of-the-art}
\newcommand{\SOTA}{\ac{SOTA}\xspace}

\acrodef{ACS}[ACS]{average cosine similarity}
\newcommand{\ACS}{\ac{ACS}\xspace}

\acrodef{ROC}[ROC]{receiver-operating characteristic}
\newcommand{\ROC}{\ac{ROC}\xspace}

\acrodef{CNN}[CNN]{convolutional neural network}
\newcommand{\CNN}{\ac{CNN}\xspace}

\acrodef{ACC}[ACC]{main task accuracy}
\newcommand{\ACC}{\ac{ACC}\xspace}

\acrodef{ASR}[ASR]{attack success rate}
\newcommand{\ASR}{\ac{ASR}\xspace}

\acrodef{DER}[DER]{defense effectiveness rating}
\newcommand{\DER}{\ac{DER}\xspace}

\acrodef{VLN}[VLN]{Vision-Language Navigation}
\newcommand{\VLN}{\ac{VLN}\xspace}

\newcommand{\der}{DER\xspace}
\newcommand{\asr}{ASR\xspace}
\newcommand{\acc}{ACC\xspace}













\acrodef{AT}[AT]{Adversarial Training}\newcommand{\AT}{\ac{AT}\xspace}

\acrodef{KL}[KL]{Kullback–Leibler}\newcommand{\kl}{\ac{KL}\xspace}

\acrodef{MDS}[MDS]{Multi-dimensional scaling}
\newcommand{\MDS}{\ac{MDS}\xspace}

\acrodef{FSR}[FSR]{Fingerprint Success Rate}
\newcommand{\FSR}{\ac{FSR}\xspace}

\makeatletter
\providecommand*{\acplural}[1]{%
  \ifAC@footnote
    \acsp{#1}%
  \else
    \acp{#1}%
  \fi}
\makeatother

\acrodef{LLM}[LLM]{Large language model}
\newcommand{\llm}{\ac{LLM}\xspace}
\newcommand{\llms}{\acp{LLM}\xspace}

\acrodef{RSC}[RSC]{Reed-Solomon Code}
\newcommand{\rsc}{\ac{RSC}\xspace}

\acrodef{pFL}[pFL]{personalized Federated Learning}
\newcommand{\pFL}{\ac{pFL}\xspace} 

\section{Introduction}
\label{sec:intro}

\begin{figure}[t]
\centering
\includegraphics[width=.9\linewidth]{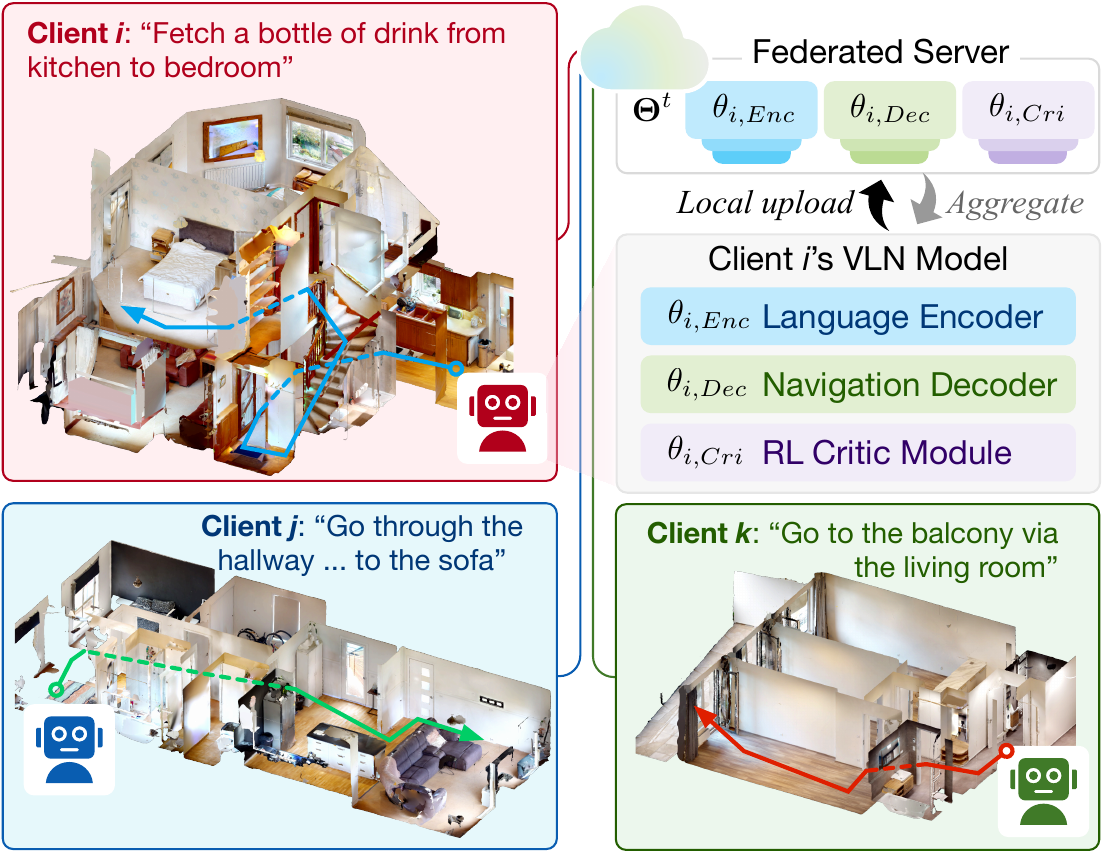}
\caption{Environmental heterogeneity across different houses for \VLN task. Each client corresponds to a distinct house with substantially different spatial layout and structural characteristics.} 
\label{fig:motivate-pfl}
\vspace{-0.25in} 
\end{figure}

\VLN has emerged as a popular and important task in embodied AI, where an agent must interpret natural language instructions and navigate through a visual environment~\cite{VLN,FedVLN}. 
Solving VLN hinges on abundant, high-quality training data because the agent must perform complex multimodal reasoning, grounding ambiguous natural language in visual observations to reach the target~\cite{FDA}.  
\VLN data often comes from \textit{private homes or offices}---navigation trajectories and human instructions that can reveal \textit{sensitive information} like house layouts, objects present, or personal habits. Fig.~\ref{fig:motivate-pfl} illustrates typical VLN scenarios and the heterogeneity across different houses for embodied AI agents executing VLN tasks. 
Indeed, conventional VLN training assumes centralizing all user data on a server, which ignores privacy concerns. Recent studies have noted that most VLN research has overlooked these real-world privacy issues~\cite{FedVLN,FedVLA}, creating a gap between academic work and practical deployment in personal spaces.

\FL offers a promising solution to this dilemma by enabling privacy-preserving collaborative training. In an FL setting, each embodied agent (or each environment) keeps its data locally and only shares model updates with a central server.
For example, FedVLN~\cite{FedVLN} was recently proposed as the first federated \VLN framework, showing that decentralized training on house-specific data can achieve navigation performance comparable to centralized training while keeping each user's data private. 
As Fig.~\ref{fig:motivate-pfl} shows, each house environment in FedVLN is treated as a client that trains a local VLN agent on its private data and periodically shares only the model parameters for aggregation. 

However, standard FL alone is not enough---a single global model cannot adequately serve every user when the data are highly heterogeneous. FedVLN relies on the basic FedAvg aggregation~\cite{mcmahan2017communication}, which struggles under VLN's \textit{severe non-IID conditions}. 
In practice, different clients' navigation data follow very different distributions, so blindly averaging their models can lead to an aggregated agent that \textit{deviates from the optimal} for any individual environment. This limitation motivates the need for personalization on top of federated training.

\begin{figure}[t]
\centering
\includegraphics[width=.98\linewidth]{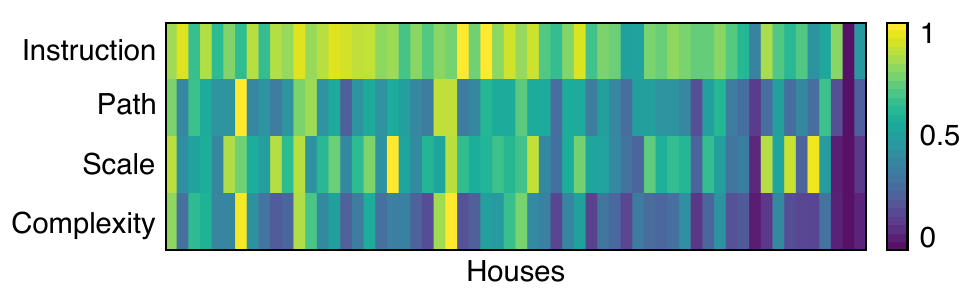}
\vspace{-0.1in}
\caption{Data heterogeneity analysis on RxR dataset~\cite{RxR}. We visualize house-level statistics along four dimensions: \textit{Instruction}, measured by the average instruction length; 
\textit{Path}, measured by the variance of {navigation trajectory lengths} within each house;
\textit{Scale}, measured by the number of rooms, indicating the house size;
and \textit{Complexity}, measured by the number of nodes in the induced navigation graph, indicating how many navigation states and decisions the agent needs to encounter. 
All statistics are normalized to [0,1] across houses.
Darker colors indicate lower values, while lighter colors indicate higher values.}
\label{fig:heatmap}
\vspace{-0.25in} 
\end{figure}

\textbf{Why \pFL for VLN?} In real-world VLN deployments, the data heterogeneity is extreme, calling for \pFL rather than a \textit{one-size-fits-all} model. Specifically, enabling \pFL for \VLN is challenging because of several unique characteristics: 
\textbf{1) Heterogeneous Environments:} Each FL client corresponds to a distinct physical environment, e.g., different houses in Room-to-Room (R2R) and Room-across-Room (RxR).
In particular, environments differ significantly in their spatial extent, layout organization, and navigation structure, resulting in induced navigation graphs with various sizes, connectivity patterns, and trajectory distributions. As illustrated in Fig.~\ref{fig:heatmap}, different clients exhibit substantial heterogeneity in their underlying navigation environments along multiple complementary dimensions, including house scale, structural complexity, and path characteristics. 
\noindent\textbf{2) Personalized Instructions:} Language instructions are inherently personal and context-specific. Different users may describe routes in unique ways---with different vocabulary, levels of detail, or referring to custom landmarks. Fig.~\ref{fig:heatmap} further highlights such linguistic heterogeneity across clients.
\textbf{3) Complex Multimodal Models:} \VLN agents typically consist of multiple modules (vision encoder, language encoder, attention-based policy network, sometimes a value estimator), which interact sequentially to produce navigation decisions. 

To address the above challenges, we propose \sys, a model structure-aware personalized federated learning framework tailored for vision-language navigation. The key insight behind \sys is that effective VLN personalization must be both structure-aware and dynamically adaptive. 
In other words, \sys treats different parts of the VLN agent differently, allowing each client to adapt the model's components to its data in a fine-grained way rather than applying coarse global updates. Building on this insight, we make the following contributions:
\begin{itemize}
    \item \textbf{Selective Module Personalization:} We develop a layer-wise adaptive personalization strategy that learns which parts of the VLN model to personalize for each client. 
    \item \textbf{Fine-Grained Parameter Fusion:} For the model parts identified as needing personalization, we propose a fine-grained parameter fusion mechanism to combine global and local knowledge.
    \item Our extensive experiments on standard VLN benchmarks, R2R~\cite{VLN} and RxR~\cite{RxR}, confirm that \sys significantly outperforms conventional federated VLN models (which lack personalization) and even approaches the performance of centrally-trained models. 
    Specifically, \sys achieves up to 7.5\% improvement in success rate, 7.8\% in normalized dynamic time warping (nDTW), highlighting its effectiveness in improving task completion as well as trajectory fidelity in \VLN. 
    Moreover, \sys achieves 1.38$\times$ faster loss convergence than baseline.
\end{itemize}

\section{Background and Motivation}
\label{sec:related}

\subsection{Vision-Language Navigation (VLN)}

\VLN~\cite{VLN} is a multimodal navigation task in which an embodied AI agent must follow natural-language instructions to reach a target location by making sequential decisions from egocentric visual observations. 
%
A defining property of VLN is its strong \textit{environment heterogeneity}. In standard benchmarks such as R2R~\cite{VLN} and RxR~\cite{RxR}, each environment (e.g., a building/scan) presents a distinct indoor layout, topology, object distribution, and navigation affordances, leading to substantially different visual and trajectory distributions across environments. 

Meanwhile, VLN instructions are naturally \textit{personalized}: different users (or annotators) describe the same route with different vocabulary, granularity, and landmark choices, and this variability is further amplified in multilingual settings such as RxR~\cite{RxR}.
These two factors jointly induce severe non-IID behavior in VLN training and deployment.

To improve generalization, prior work has explored diverse modeling and training strategies, including imitation and reinforcement learning, structured reasoning via graph-based representations~\cite{Dualscale}, long-term memory for accumulating navigation context~\cite{Iterative}, and data augmentation / fine-tuning with synthesized trajectory-instruction 
pairs~\cite{ENVDROP}.
Recent studies further consider pre-exploration for better adaptation to unseen environments~\cite{Lookahead} and leverage LLMs for explicit planning and reasoning (e.g., NavGPT~\cite{NavGPT}).
However, these methods largely assume \emph{centralized} access to training data, which is often unrealistic in practice because VLN trajectories and instructions are collected in private indoor spaces (e.g., homes, offices, and labs), making raw data sharing undesirable. 

\subsection{Federated VLN Formulation}

FedVLN~\cite{FedVLN}, recently proposed as the first federated \VLN framework, trains the trajectory model, language encoder, multimodal decision module, and speaker model across clients using FedAvg~\cite{mcmahan2017communication}, and further introduces a federated pre-exploration phase for adaptation after deployment. 
%
A VLN agent deployed in a house environment $i$ acts as a client in FL.
In each local round, the agent follows a set of human linguistic instructions $\mathcal{I}=\{I_1,I_2,\ldots\}$, where each instruction $I\in\mathcal{I}$ is a sequence of tokens
$I=\langle w_1,w_2,\ldots,w_m\rangle$ ($m$ is the instruction length and $w_j$ is a word token).
For each instruction, the agent executes a navigation episode and produces a trajectory
$\tau=\langle o_0,a_0,\ldots,o_T\rangle$ by interacting with the environment, terminating upon issuing a \texttt{STOP} action.
Each local dataset $\mathcal{D}_i$ therefore consists of instruction–trajectory pairs $(I,\tau)$.
%

The local VLN model contains three modules: a language instruction encoder $\theta_{Enc}^{t}$, a navigation decoder $\theta_{Dec}^{t}$, and a critic module $\theta_{Cri}^{t}$ for RL learning.
At communication round $t$, client $i$ receives the global parameters
$\boldsymbol{\Theta}^{t}=\{\theta_{Enc}^{t},\theta_{Dec}^{t},\theta_{Cri}^{t}\}$,
and performs local optimization on its private dataset $\mathcal{D}_i$ (paired instruction--trajectory samples).

After local training, the updated parameters $\boldsymbol{\Theta}_i^{(t+1)}$ are uploaded to the server.
The server aggregates updates from a subset of participating clients $\mathcal{S}_t \subseteq \{1,\ldots,N\}$ to construct the next-round global model.
This federated optimization proceeds iteratively across rounds.
The objective of vanilla federated VLN is to collaboratively learn a navigation policy that can operate across diverse embodied environments without sharing raw data.

\begin{figure}[t]
    \centering
    \includegraphics[width=.9\linewidth]{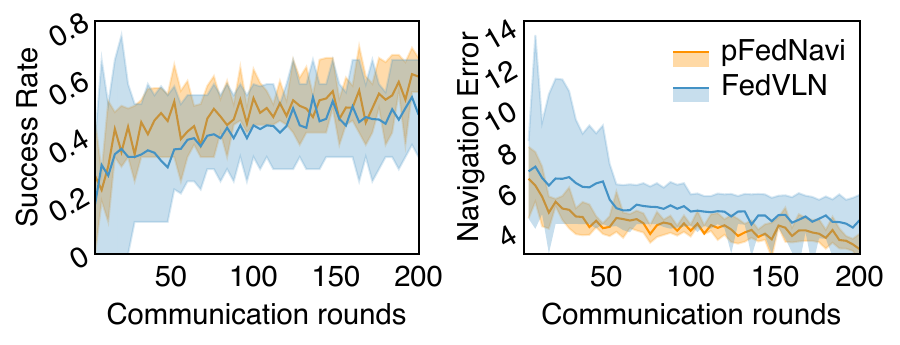}
    \vspace{-0.1in}
    \caption{Success rate and navigation error comparison on R2R dataset. The line and the shadow mean the average and variance of performance across clients.}
    \label{fig:improvement-pfl}
    \vspace{-0.25in} 
\end{figure}

\subsection{Motivating \pFL for VLN}
While FL protects privacy and enables knowledge sharing, \emph{vanilla} federated VLN (e.g., FedVLN~\cite{FedVLN}) that aggregates all client models into a single global model often \textit{struggles} under VLN’s severe data heterogeneity.
This has motivated \pFL, which aims to learn client-adaptive models rather than enforcing a \textit{one-size-fits-all} solution.
Fig.~\ref{fig:improvement-pfl} shows that FedAvg-based VLN exhibits large performance variance across \FL clients, demonstrating that a single model cannot sufficiently serve all environments. 

Existing \pFL methods include model-splitting approaches such as FedCP~\cite{Fedcp}, which separate globally shared and client-specific components (e.g., backbone \textit{vs.}\ head), as well as full-model adaptation approaches such as Per-FedAvg-style personalization~\cite{perFed}. 
However, \VLN introduces \textit{additional difficulties} due to multimodal grounding and tightly coupled encoder-decoder architectures, where indiscriminate personalization can degrade language grounding or action prediction.

\begin{figure}[t]
\centering
\includegraphics[width=0.74\linewidth]{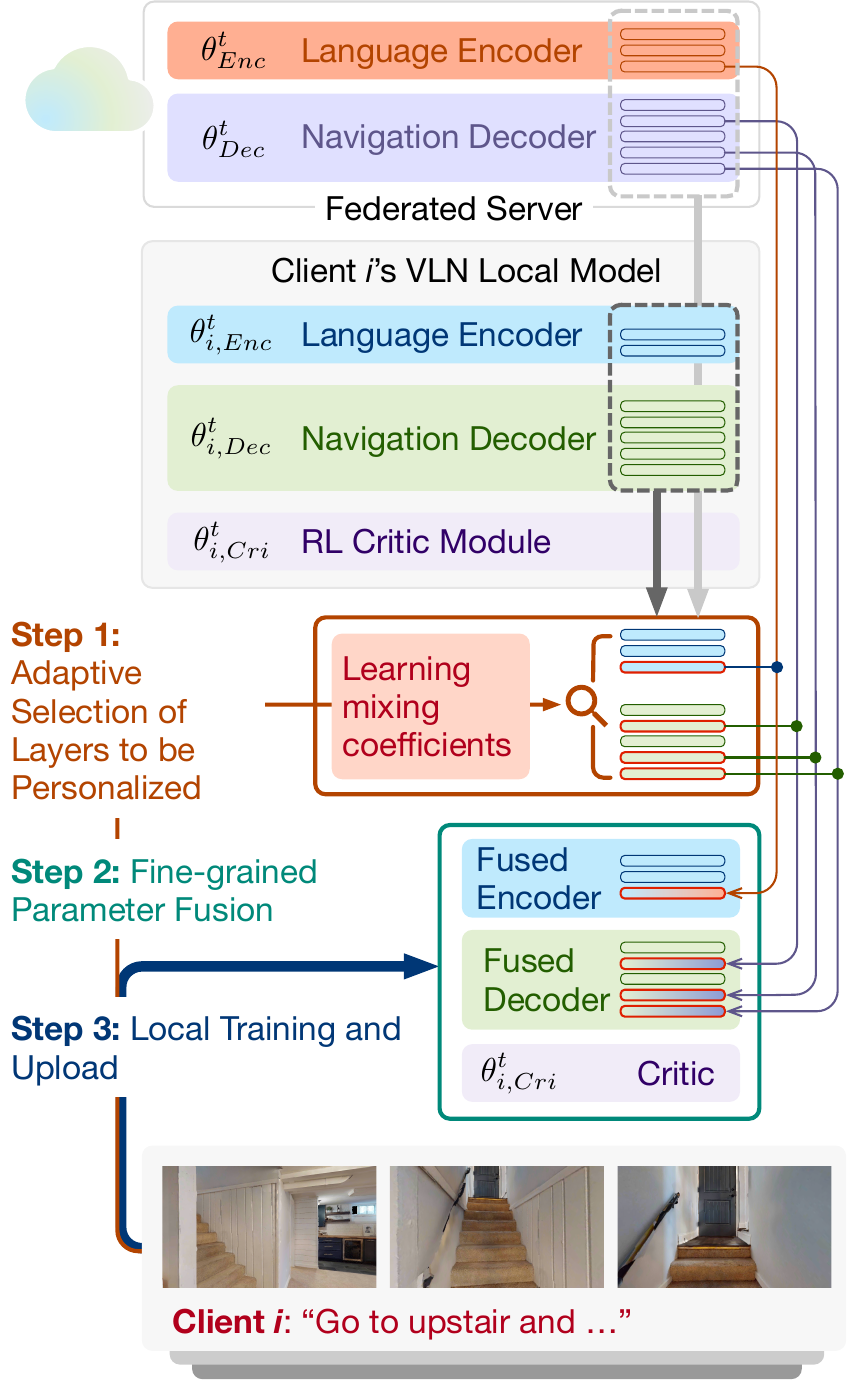}
\caption{\sys's workflow. \sys operates in three stages:
(1) adaptive personalized layer selection;
(2) fine-grained parameter fusion;
and (3) local training with federated aggregation.}
\vspace{-0.25in}
\label{fig:framework}
\end{figure}

\section{The Design of \sys}
\label{sec:design}

\subsection{Overview}

Enabling pFL for VLN is fundamentally challenging due to the joint effects of multimodal inputs, tightly coupled encoder-decoder structures, and strong environment-induced heterogeneity across clients.   
Building on the key insight that effective VLN personalization must be \emph{structure-aware} and \emph{dynamically adaptive}, we design \sys in Fig.~\ref{fig:framework}. 

\textbf{Design Objectives:} 1) Client-adaptive navigation without sacrificing shared generalization; 2) Structure-aware personalization that preserves grounding; and 3) Stable and data-efficient personalization under severe non-IID.

\textbf{Key Challenges:} 
These objectives translate into three key challenges that directly motivate our design components:

\begin{itemize}[leftmargin=*,topsep=0.2em,itemsep=0.2em]
    \item \textbf{Non-IID drift makes naive aggregation suboptimal.}
    With extreme cross-environment and cross-instruction heterogeneity, FedAvg-style averaging can produce a global model that is not optimal for any particular client, and local updates can be inconsistent across rounds. 

    \item \textbf{VLN module coupling makes indiscriminate personalization harmful.}
    Unlike standard pFL settings, blindly personalizing arbitrary layers can easily disrupt language grounding or action prediction because VLN decisions rely on tightly coupled encoder-decoder interactions. 

    \item \textbf{``Which layers to personalize'' is client- and round-dependent.}
    Different environments and instruction styles stress different model components; thus, a fixed model split (e.g., backbone and head) is often insufficient, and we need to adaptively \textit{recognize} and \textit{personalize} layers in the global model with clients' local information.
\end{itemize}

\textbf{A Bird's Eye View of \sys:} 
\textbf{Step \circled{1} Adaptive layer selection:} We first identify model components that require personalization, including the encoder-decoder projection layer and selected decoder layers, using adaptive layer-wise mixing. 
\textbf{Step \circled{2} Fine-grained parameter fusion:} We apply fine-grained parameter fusion to the selected components to balance global knowledge and local specialization.
\textbf{Step \circled{3} Local training \& upload:} We initialize the local model using the above steps and train on the private data. 
\subsection{Formulating \pFL for \VLN}

\noindent\textbf{Personalized federated objective.}
Let there be $N$ clients, where Client $i$ corresponds to a VLN agent deployed in environment $i$ with private dataset $\mathcal{D}_i$ (paired instruction--trajectory samples). 
Unlike vanilla FL that optimizes a single shared model, pFL for VLN aims to learn a set of client-specific models $\{\boldsymbol{\Theta}_i\}_{i=1}^N$:
$\min_{\{\boldsymbol{\Theta}_i\}} \sum_{i=1}^{N}
\mathbb{E}_{(I,\tau)\sim \mathcal{D}_i}
\Big[
\mathcal{L}_{\mathrm{VLN}}(I,\tau; \boldsymbol{\Theta}_i)
\Big],$
where $\mathcal{L}_{\mathrm{VLN}}$ denotes the standard VLN training objective (e.g., imitation loss and, when used, RL loss/critic terms).
To encourage knowledge sharing, we maintain a \emph{global reference} model $\boldsymbol{\Theta}$ on the server, constructed by aggregating client updates on globally shared parameters.

\noindent\textbf{Why structure-aware personalization.}
VLN models differ from standard unimodal networks because language understanding and action prediction are tightly coupled through structured modules.
Different components therefore exhibit different degrees of environment sensitivity.
Accordingly, \sys treats the VLN model as three modules and personalizes them differently:

\begin{itemize}[leftmargin=*,topsep=0.2em,itemsep=0.2em]
    \item \textbf{Encoder.}
    We follow the standard VLN encoder~\cite{VLN}, consisting of (i) an embedding layer, (ii) an instruction BiLSTM encoder, and (iii) an encoder-to-decoder projection. 
    
    The embedding and BiLSTM mainly capture general linguistic regularities and are thus globally shared, while the projection 
    directly mediates language grounding into navigation intent and is more environment-sensitive.
    Therefore, \sys always includes the encoder-decoder projection layer in personalization. 

    \item \textbf{Decoder.}
    We adopt the standard attention-based decoder used in prior VLN agents~\cite{VLN,ENVDROP}, which contains multiple functional components (action embedding, visual attention, recurrent state update, instruction attention, candidate scoring).
    Since these components have heterogeneous sensitivity to local environments, \sys does \emph{not} predefine personalized decoder layers; instead, it adaptively selects them for each client and each round (Sec.~\ref{sec:layer_selection}).

    \item \textbf{Critic.}
    The critic estimates value functions that depend heavily on environment-specific dynamics and reward landscapes.
    To avoid destabilizing training via cross-client averaging, \sys keeps the critic \emph{local} and does not force it to match the global reference.
\end{itemize}

\subsection{Adaptive Layer Selection for Personalization}  
\label{sec:layer_selection} 

Given the received global model and the client's previous local model, we want to decide \emph{which} decoder layers should be personalized for client $i$ at round $t$.

Therefore, instead of predefining personalized layers, we adopt a layer-wise adaptive fusion mechanism to automatically identify decoder components that benefit from personalization. 
Specifically, we associate each decoder component (layer) $\ell \in \text{L}_{\mathrm{dec}}$ with a learnable mixing coefficient $\alpha_{i,\ell}^{t}\in[0,1]$ for client $i$ at round $t$.
The fused parameters are defined as
$
\hat\theta_{i,\ell}^{t+1,0}
=
\bigl(1-\alpha_{i,\ell}^{t}\bigr)\,\theta_{\ell}^{t}
+
\alpha_{i,\ell}^{t}\,\theta_{i,\ell}^{t}, 
\label{eq:dec_layer_fuse}
$
where $\theta_{\ell}^{t}$ denotes the globally aggregated parameters of decoder layer $\ell$ at round $t$, and $\theta_{i,\ell}^{t}$ is the corresponding previous local parameters.

We learn $\boldsymbol{\alpha}_{i}^{t}=\{\alpha_{i,\ell}^{t}\}_{\ell\in\text{L}_{\mathrm{dec}}}$ by minimizing the teacher-forcing imitation loss on a small local batch set $\mathcal{B}_i \subset \mathcal{D}_i$:
$
\boldsymbol{\alpha}_{i}^{t}
\leftarrow
\arg\min_{\boldsymbol{\alpha}}
\ \mathcal{L}_{\mathrm{IL}}\!\left(\mathcal{B}_i;\ \hat\theta_{i,\mathrm{Dec}}^{t+1,0}(\boldsymbol{\alpha})\right).$ 
Layers selected for personalization are then determined by a thresholding rule: 
\begin{equation}
\text{Personalized\_layer}_{i}^{t}
=
\left\{\ell\in\text{L}_{\mathrm{dec}}\ \big|\ \alpha_{i,\ell}^{t}\ge \delta \right\}.
\label{eq:layer_select}
\end{equation}
Layers with a small $\alpha_{i,\ell}^{t}$ are treated as globally shared and directly inherited from the global model.
Here, $\delta$ controls how strongly a layer must prefer local parameters to be considered personalized. 
We set $\delta=0.6$ rather than 0.5 to avoid marginal or ambiguous personalization decisions.

\subsection{Fine-grained Parameter Fusion}

After identifying personalized components, we perform \emph{fine-grained parameter fusion} to initialize the client’s personalized model before local training.

For client $i$ at communication round $t$, let $\mathcal{K}_{i}^{t}$ denote the set of all layers selected for personalization, which includes:
(i) encoder-decoder projection layer, and
(ii) $\text{Personalized\_layer}_{i}^{t}$ identified in Equation~(\ref{eq:layer_select}).
For each layer $l \in \mathcal{K}_{i}^{t}$, we construct the personalized initialization via element-wise interpolation between the global and previous local parameters:
\begin{equation}
    \theta^{t+1,0}_{i,l}
    =
    \theta^{t}_{i,l}
    +
    W^{t}_{i,l} \odot
    \bigl(
    \theta^{t}_{l}
    -
    \theta^{t}_{i,l}
    \bigr),
    \label{eq:model-fusion}
\end{equation}
where $\theta^{t}_{l}$ denotes the globally aggregated layer parameter,
$\theta^{t}_{i,p}$ is the corresponding parameter from the previous local model, and
$W^{t}_{i,l}\in[0,1]$ is a learnable fusion weight controlling the contribution of global knowledge.
This design enables flexible personalization while constraining the additional cost to preserve scalability, as fusion weights are introduced only for a small subset of decoder components.

Parameters not selected for personalization are directly inherited from the global model:
\begin{equation}
    \theta^{t+1,0}_{i,p} \leftarrow \theta^{t}_{p},
    \quad \forall p \notin \mathcal{K}_{i}^{t}.
    \label{eq:nonPL_init}
\end{equation}

The critic module is kept local and initialized as
\begin{equation}
    \theta^{t+1,0}_{i,\mathrm{Cri}} \leftarrow \theta^{t}_{i,\mathrm{Cri}},
    \label{eq:critic_init}
\end{equation}
since it primarily estimates value functions conditioned on environment-specific dynamics.
The fusion weights $\mathbf{W}_{i}^{t} = \{W^{t}_{i,l}\}_{l\in\mathcal{K}_{i}^{t}}$ are optimized by minimizing the supervised imitation loss under teacher forcing:
$\min_{\mathbf{W}_{i}^{t}} \mathcal{L}_{\mathrm{IL}}\!\left(\mathcal{D}_i;\,\boldsymbol{\Theta}_{i}^{t+1,0}(\mathbf{W}_{i}^{t})\right)$,
which adaptively balances global generalization and local specialization for each personalized parameter.

\subsection{Local Training and Upload}

After learning the fusion weight, we calculate the initialized personalized model $\boldsymbol{\Theta}_{i}^{t+1,0}$ by Equations~(\ref{eq:model-fusion}, \ref{eq:nonPL_init}, \ref{eq:critic_init}) and then train on the client’s local dataset $\mathcal{D}_i$ using the standard VLN objective, which combines imitation loss (IL) and RL loss~\cite{ENVDROP}.
After local training, the updated model $\boldsymbol{\Theta}_{i}^{t+1}$ is uploaded to the server for aggregation:
$
\boldsymbol{\Theta}^{t+1}
=
\sum_{i\in\mathcal{S}_t}
\frac{|\mathcal{D}_i|}{\sum_{j\in\mathcal{S}_t}|\mathcal{D}_j|}
\,
\boldsymbol{\Theta}_{i}^{t+1},
$
where $\mathcal{S}_t$ denotes the set of participating clients at round $t$ and $|\mathcal{D}_i|$ is the size of client $i$’s local dataset.
The aggregated global model $\boldsymbol{\Theta}^{t+1}$ is then broadcast to clients for the next communication round.

 \section{Evaluation}
\label{sec:evaluation}
\subsection{Setup}

\textbf{Dataset and Model:}
We evaluate \sys on two widely used VLN datasets: Room-to-Room (R2R)~\cite{VLN} and Room-across-Room (RxR)~\cite{RxR}, which are both constructed from the Matterport3D dataset.
R2R contains 7K navigation trajectories paired with human-written English instructions.
RxR~\cite{RxR} is a larger multilingual dataset with more trajectories and denser linguistic grounding, where each word is time-aligned to annotator viewpoints.
We follow the standard data preprocessing and environment configurations used in prior VLN work~\cite{FedVLN}.
We adopt the VLN agent architecture consistent with prior studies~\cite{VLN} and evaluate it with two widely used alternative pretrained visual feature extractors, ResNet-152 and CLIP.
The navigation instructions are encoded using an LSTM-based language encoder, and an attention-based LSTM decoder attends to both textual and visual features to predict actions at each step.
The critic module is implemented to estimate state values during reinforcement learning.

\textbf{FL Setup:}
We regard each agent in a building environment as an FL client.
All experiments are conducted on a single machine with two NVIDIA RTX 3090 GPUs bridged with NVLink.
We reuse the federated learning hyperparameters, such as the number of training rounds, the number of local epochs, and both the local and global learning rates, from FedVLN~\cite{FedVLN}.
Specifically, at each communication round, a subset of clients is randomly sampled with a participation rate of $S_r=0.2$.
Each participating client performs local training for 5 epochs, and all models are trained until convergence.
For personalized layer selection, the mixing coefficients $\boldsymbol{\alpha}$ are optimized using a learning rate of $\alpha_{\mathrm{lr}}=0.1$ for $S_{\alpha}=2$ steps.
We select personalized decoder layers using a threshold $\delta=0.6$.
For fine-grained fusion, the fusion weights are updated with a learning rate of $\eta=0.1$ for $S_{W}=1$ step per round, and are fully optimized until convergence only in the second round, after which they are lightly fine-tuned.

\begin{table}[t]
\centering
\caption{Evaluation Results on R2R and RxR Datasets (all metrics except NE in \%). ResNet and CLIP indicate the visual feature extractors used by VLN agent, with the model architecture remaining identical. $\uparrow$ indicates higher values correspond to better performance, whereas $\downarrow$ indicates lower values are better. \textit{Please note that EnvDrop~\cite{ENVDROP} is a centralized learning baseline.}}
\label{tab:r2r_results}
\setlength{\tabcolsep}{3pt}
\resizebox{\columnwidth}{!}{%
\begin{tabular}{llcccccc}
\toprule
 & \textbf{Method on R2R} 
& \textbf{SR $\uparrow$} & \textbf{SPL $\uparrow$} & \textbf{OSR $\uparrow$} 
& \textbf{CLS $\uparrow$} & \textbf{nDTW $\uparrow$} & \textbf{NE $\downarrow$} \\
\midrule
\multirow{3}{*}{ResNet} 
& EnvDrop~\cite{ENVDROP} & \textbf{56.7} & \textbf{54.3} & 63.9 & \textbf{67.2} & \textbf{56.0} & \textbf{4.55} \\
& FedVLN~\cite{FedVLN}   & 50.7 & 47.3 & 60.3 & 63.4 & 50.8 & 5.37 \\
\rowcolor{gray!20}
\cellcolor{white}& \sys                  & 54.5 & 51.7 & \textbf{65.2} & 66.4 & 54.8 & 4.86 \\
\midrule
\multirow{3}{*}{CLIP} 
& EnvDrop~\cite{ENVDROP} & \textbf{61.5} & 56.5 & \textbf{69.8} & 67.3 & 55.4 & 3.94 \\
& FedVLN~\cite{FedVLN}    & 59.5 & 56.3 & 65.1 & 67.5 & 55.1 & 4.24 \\
\rowcolor{gray!20}
\cellcolor{white}& \sys                   & 60.7 & \textbf{57.4} & 67.3 & \textbf{68.6} & \textbf{57.1} & \textbf{3.65} \\
\toprule
 & \textbf{Method on RxR} 
& \textbf{SR $\uparrow$} & \textbf{SPL $\uparrow$} & \textbf{OSR $\uparrow$} 
& \textbf{CLS $\uparrow$} & \textbf{nDTW $\uparrow$} & \textbf{NE $\downarrow$} \\
\midrule
\multirow{3}{*}{ResNet} 
& EnvDrop~\cite{ENVDROP} & \textbf{41.3} & 36.6 & 50.4 & 57.2 & \textbf{53.6} & \textbf{8.03} \\
& FedVLN~\cite{FedVLN}   & 39.9 & 33.1 & 47.8 & 55.5 & 50.7 & 8.62 \\
\rowcolor{gray!20}
\cellcolor{white}& \sys                  & 40.1 & \textbf{36.7} & \textbf{51.2} & \textbf{57.3} & 52.9 & 8.10 \\
\midrule
\multirow{3}{*}{CLIP} 
& EnvDrop~\cite{ENVDROP} & \textbf{47.7} & \textbf{43.4} & 56.1 & \textbf{61.2} & 56.8 & \textbf{6.53} \\
& FedVLN~\cite{FedVLN}    & 43.0 & 39.4 & 51.5 & 58.7 & 54.8 & 7.37 \\
\rowcolor{gray!20}
\cellcolor{white}& \sys                   & 46.1 & 41.2 & \textbf{56.3} & 59.8 & \textbf{56.9} & 6.91 \\
\bottomrule
\end{tabular}
}
\vspace{-15pt}
\end{table}

\textbf{Baselines:} 
We compare our method with both centralized and federated learning approaches.
For centralized training, we use \textbf{EnvDrop}~\cite{ENVDROP}, which employs a BiLSTM-based language encoder, an attentive LSTM decoder for action prediction.
We evaluate EnvDrop under two visual feature extraction settings, using either pretrained ResNet-based features or CLIP-based features. 
For federated learning, we evaluate \textbf{FedVLN}~\cite{FedVLN}, the first work that applies FedAvg~\cite{mcmahan2017communication} to the VLN setting.

\textbf{Metrics:}
To comprehensively assess navigation ability, we consider two categories of evaluation metrics: 1) goal-reaching performance and 2) trajectory fidelity. 
1) \textbf{Success Rate (SR)} computes the proportion of episodes in which the agent issues a \texttt{stop} action within 3 meters of the target. 
\textbf{Success weighted by Path Length (SPL)} extends SR by incorporating path optimality, rewarding agents that reach the goal with shorter and more efficient paths. 
\textbf{Oracle Success Rate (OSR)}
reflects the agent’s ability to at least visit a promising region even if it fails to stop correctly. 
\textbf{Navigation Error (NE)} reports the average distance between the agent’s final position and the goal location, providing a continuous measure of goal accuracy.
2) \textbf{Coverage weighted by Length Score (CLS)} jointly considers spatial coverage and trajectory length to evaluate how well the agent explores relevant regions. 
\textbf{Normalized Dynamic Time Warping (nDTW)} aligns the predicted and reference trajectories to quantify their similarity. 
Together, these metrics assess high-level task completion as well as low-level behavioral fidelity, both of which are essential for VLN evaluation. Notably, we use the average client-side value of each metric for our \sys method and compare it with the corresponding global metric value.


\begin{figure}[t]
  \centering
    \includegraphics[width=\linewidth]{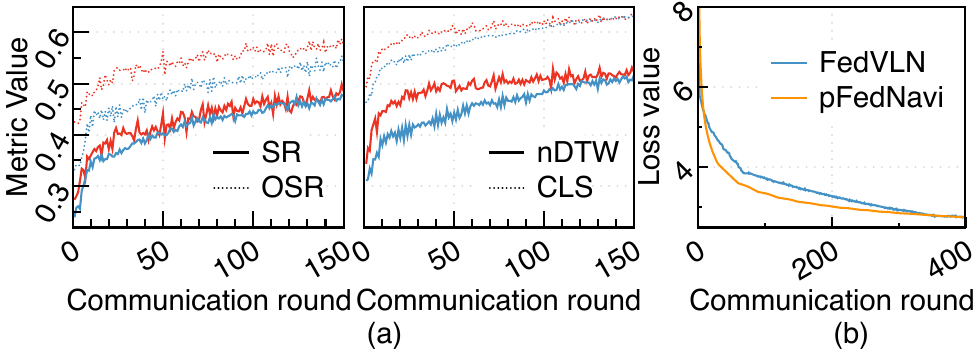}
    \vspace{-0.25in}
  \caption{Comparison between \sys and FedVLN on R2R dataset using ResNet-152 visual features. (a) The performance curves of various metrics (e.g., SR, OSR, nDTW, and CLS), and (b) loss convergence over communication rounds (\textcolor[HTML]{D84531}{red}: \sys, \textcolor[HTML]{588FBF}{blue}: FedVLN).}
  \label{fig:convergence}
  \vspace{-0.2in}
\end{figure}


\subsection{The Effectiveness of \sys}
Table~\ref{tab:r2r_results} summarizes the performance comparison among centralized training, FedAvg-based federated learning, and our personalized federated approach on both R2R and RxR datasets.
Overall, \sys consistently outperforms FedVLN across all metrics under both ResNet-152 and CLIP visual representations, demonstrating the effectiveness of personalization for VLN in the federated setting.
Compared with FedVLN, \sys's improvement is particularly evident on metrics that reflect trajectory quality and goal grounding.
These gains indicate that personalized federated learning enables each client to better adapt navigation policies to its local environment, alleviating the bias introduced by naive parameter averaging under heterogeneous VLN data.

Besides, we set the client participation rate to $S_r = 1$ for clearer visualization of learning dynamics.
Fig.~\ref{fig:convergence}(a) compares the learning dynamics of \sys and FedVLN on the R2R dataset with ResNet features.
Across all metrics, \sys achieves faster performance improvement.
Notably, the more rapid gains on OSR and nDTW indicate that personalized parameter fusion helps preserve environment-specific navigation patterns and instruction grounding, whereas FedVLN improves more slowly due to averaging across heterogeneous clients.

\subsection{Convergence and Overhead Analysis \& Ablation Study}
Fig.~\ref{fig:convergence}(b) shows the convergence performance compared with FedVLN, demonstrating that \sys converges to the $\text{target loss}=3.0$ in 208 communication rounds, whereas FedVLN requires 288 communication rounds, resulting in a 27.8\% improvement in efficiency (1.38$\times$ faster convergence). Despite this short-term fluctuation, \sys converges substantially faster thereafter.
Since the critic module is always kept local and only the encoder and decoder modules are transmitted each round, our communication overhead is lower than FedVLN.
In wall-clock time, our approach costs $3.6$ min per round compared with $2.2$ min for FedVLN. 

We evaluate three variants to analyze the impact of personalized layer selection: 1) All layers, where all model parameters conduct parameter fusion for each client; 2) No layer, where all parameters are globally shared (FedVLN); and 3) \sys. 
Table~\ref{tab:ablation} shows that personalizing all layers leads to a clear performance degradation across all metrics. 
This indicates that fully localizing the entire VLN model is ineffective under federated settings, as each client typically has limited data and cannot reliably learn both general navigation knowledge and environment-specific behaviors.
In addition, personalizing all layers incurs substantially higher computational and storage overhead.
In contrast, \sys selectively personalizes structure-sensitive components, achieving superior performance while significantly reducing training cost and model storage.

\begin{table}[t]
\centering
\caption{Evaluation results for various personalized layer selection strategies. 
}
\vspace{-0.05in}
\label{tab:ablation}
\setlength{\tabcolsep}{6pt} 
\resizebox{\columnwidth}{!}{%
\begin{tabular}{lcccccc}
\toprule
& \textbf{SR $\uparrow$} & \textbf{SPL $\uparrow$} & \textbf{OSR $\uparrow$} 
& \textbf{CLS $\uparrow$} & \textbf{nDTW $\uparrow$} & \textbf{NE $\downarrow$} \\
\midrule
All layers & 32.7 & 30.1 & 41.0 & 42.5 & 45.2 & 8.52\\
No layer   & 39.9 & 33.1 & 47.8 & 55.5 & 50.7 & 8.62 \\
\rowcolor{gray!20}
\sys       & \textbf{40.1} & \textbf{36.7} & \textbf{51.2} & \textbf{57.3} & \textbf{52.9} & \textbf{8.10} \\
\bottomrule

\end{tabular}
}
\vspace{-0.15in}
\end{table}

\section{Conclusion} 
\label{sec:conclusion}

This paper studies pFL for VLN under realistic heterogeneity, where each client corresponds to a distinct indoor environment with unique spatial structure and personalized instructions. We propose \sys, a structure-aware personalized FL framework that adaptively selects components for personalization via layer-wise mixing and initializes client models through fine-grained parameter fusion, effectively balancing global generalization and local adaptation. Experiments on R2R and RxR demonstrate that \sys consistently outperforms FedVLN, achieving up to 7.5\% improvement in success rate and 7.8\% improvement in normalized dynamic time warping, while converging 1.38$\times$ faster.

\bibliographystyle{IEEEtran}
\bibliography{IEEEabrv,main}

\end{document}